\documentclass{article}

\usepackage[final]{corl_2026}

\usepackage[utf8]{inputenc}
\usepackage[T1]{fontenc}
\usepackage{url}
\usepackage{booktabs}
\usepackage{amsfonts}
\usepackage{amsmath}
\usepackage{microtype}
\usepackage{xcolor}
\usepackage{graphicx}
\usepackage{subcaption}
\usepackage{wrapfig}
\usepackage{multirow}

\newcommand{\eg}{e.g.\@}
\definecolor{impgreen}{RGB}{34,139,34}
\newcommand{\up}[1]{\textsubscript{\textcolor{impgreen}{$\uparrow$#1}}}
\newcommand{\blfootnote}[1]{\begingroup\renewcommand\thefootnote{}\footnote{#1}\addtocounter{footnote}{-1}\endgroup}

\title{Semantic Anchoring for\\ Robotic Action Representations}

\author{
  Yuan Xu$^{*,1,2}$,~
  Youheng Shi$^{*,2}$,~
  Chengyang Li$^{4,5}$,~
  Wentao Zhu$^{\dagger,4}$,~
  Yizhou Wang$^{\dagger,1,3}$ \\[6pt]
  $^{1}$Center on Frontiers of Computing Studies, School of Computer Science, Peking University\\
  $^{2}$AI Innovation Center, School of Computer Science, Peking University\\
  $^{3}$State Key Laboratory of General Artificial Intelligence, Peking University\\[2pt]
  $^{4}$Eastern Institute of Technology, Ningbo\quad
  $^{5}$Shanghai Jiao Tong University
}

\providecommand{\Dset}{\mathcal{D}}

\providecommand{\obs}{\mathbf{o}}
\providecommand{\act}{\mathbf{a}}
\providecommand{\instr}{\ell}
\providecommand{\ploss}{c}

\providecommand{\vla}{f_{\theta}}
\providecommand{\layer}{k}
\providecommand{\hraw}[1][\layer]{\mathbf{h}^{(#1)}}
\providecommand{\hpool}[1][\layer]{\bar{\mathbf{h}}^{(#1)}}
\providecommand{\hrec}{\mathbf{h}_{\mathrm{rec}}}
\providecommand{\qpool}{\mathbf{q}}

\providecommand{\genc}{g_{\text{enc}}}
\providecommand{\gprobe}{g_{\text{probe}}}
\providecommand{\etext}{\mathbf{e}}

\providecommand{\probeA}{\mathcal{T}_a}
\providecommand{\probeT}{\mathcal{T}_t}
\providecommand{\zproj}{\hat{\mathbf{z}}}

\providecommand{\Esh}{E_{s}}
\providecommand{\Epr}{E_{p}}
\providecommand{\Dec}{\mathrm{Dec}}
\providecommand{\zs}{\mathbf{z}^{s}}
\providecommand{\zp}{\mathbf{z}^{p}}

\providecommand{\Lact}{\mathcal{L}_{\text{action}}}
\providecommand{\Lalign}{\mathcal{L}_{\text{align}}}
\providecommand{\Lprobe}{\mathcal{L}_{\text{probe}}}
\providecommand{\Lrecon}{\mathcal{L}_{\text{recon}}}
\providecommand{\Ldiff}{\mathcal{L}_{\text{diff}}}
\providecommand{\Ltotal}{\mathcal{L}_{\text{total}}}

\providecommand{\bsz}{B}
\providecommand{\dproj}{d_{p}}
\providecommand{\dsub}{d_{s}}
\providecommand{\temp}{\tau}
\providecommand{\lamA}{\lambda_{\text{align}}}
\providecommand{\lamR}{\lambda_{r}}
\providecommand{\lamD}{\lambda_{d}}
\providecommand{\nact}{T}

\providecommand{\cossim}{\mathrm{sim}}

\begin{document}

\maketitle
\blfootnote{$^*$Equal contribution. $^\dagger$Corresponding authors.}

\vspace{-\intextsep}
\begin{figure}[h]
    \centering
    \includegraphics[width=\textwidth]{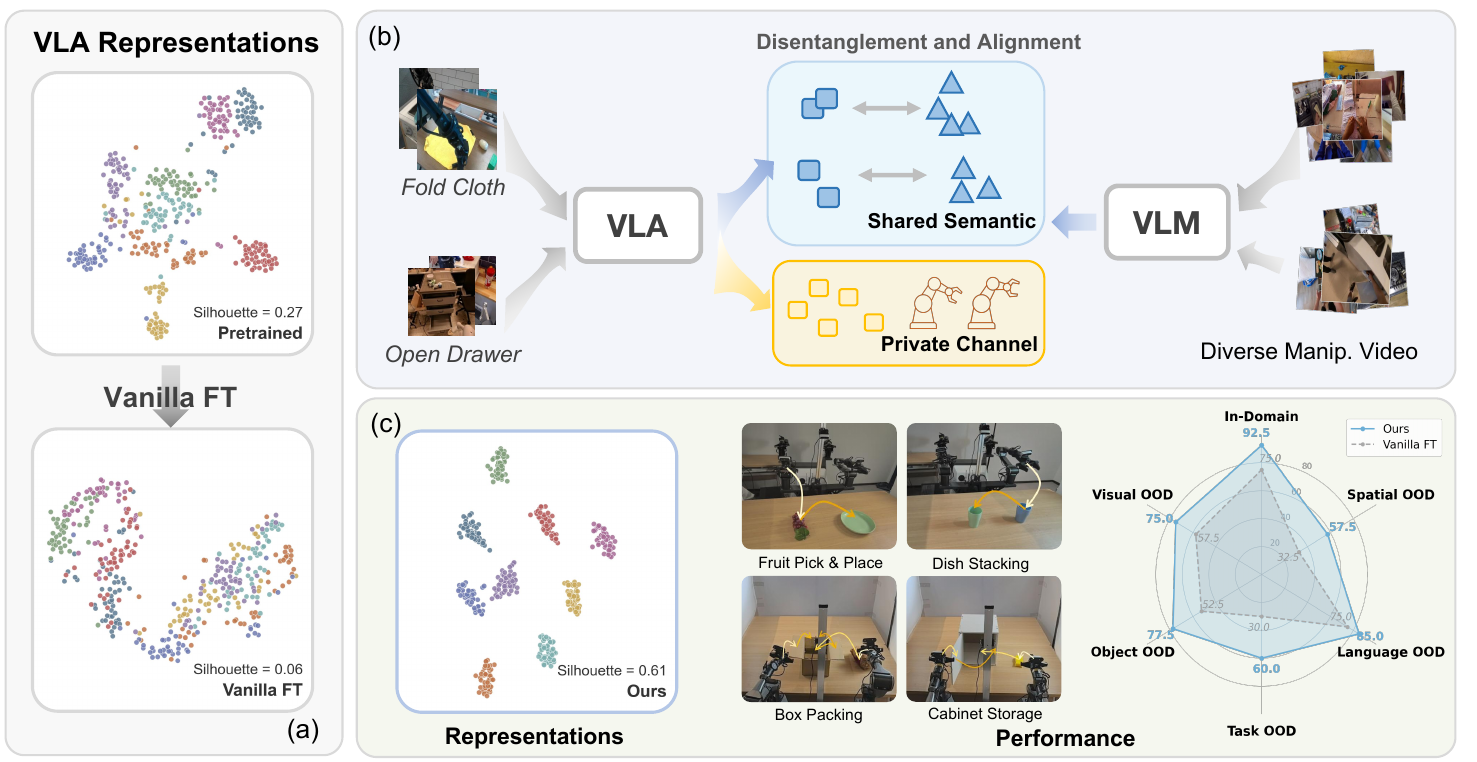}
    \caption{%
    \textbf{Semantic anchoring for robotic action representations.}
    \textbf{(a)}~Vanilla action-only fine-tuning degrades the semantic structure inherited from pretraining: t-SNE of $\pi_0$'s layer-10 features on LIBERO-Goal~\citep{liu2023libero}, colored by task.
    \textbf{(b)}~Our method anchors action representations to a semantic manifold during training via contrastive alignment with shared/private decomposition.
    \textbf{(c)}~Repaired representations recover semantic structure and consistently improve both in-distribution and OOD performance on a real bimanual platform.}
    \label{fig:teaser}
\end{figure}

\begin{abstract}
Pretrained Vision-Language-Action (VLA) models inherit rich semantic representations from Vision-Language Models, yet this structure can erode during downstream fine-tuning on limited robot demonstrations. A fundamental question therefore arises: \emph{what constitutes a good action representation?} Inspired by the mirror neuron theory's insight that observation and execution share an intention-level encoding, we examine whether a robot's action representations preserve the semantic structure captured by pretrained encoders. Systematic probing suggests that this structure erodes during fine-tuning and that its quality is associated with both task success and out-of-distribution generalization. We further introduce a method that anchors action representations to a semantic manifold during training while decomposing representations into a shared semantic channel and a private channel, all discarded at inference, leaving the deployed model unchanged. Validated on different VLA backbones across simulation and real-world benchmarks, our method yields up to $+18.7\%$ on real-world in-distribution tasks and $+21.5\%$ on out-of-distribution generalization. Project page is available at \url{https://xy02-05.github.io/SemanticMN}.

\end{abstract}


\section{Introduction}
\label{sec:intro}

Vision-Language-Action (VLA) models build generalist robot policies by fine-tuning pretrained Vision-Language Models (VLM) on robot demonstrations, inheriting a richly structured representation space that, in principle, enables generalization across novel objects, scenes, and instructions~\citep{zitkovich2023rt, kim2024openvla, black2024pi0}.
In practice, however, downstream fine-tuning of pretrained VLAs relies on demonstration data that are orders of magnitude smaller and narrower than the VLM's pretraining corpus, and action-only supervision can drive the model toward shortcut mappings that degrade the inherited semantic structure~\citep{geirhos2020shortcut, li2025shortcutrobot, gao2025dontblind}: Fig.~\ref{fig:teaser}a shows that the clear task-level clustering present in pretrained representations collapses after fine-tuning.
Existing efforts mitigate this by scaling up data~\citep{intelligence2025pi05, bjorck2025gr00t} or redesigning architectures~\citep{ghosh2024octo, qu2025spatialvla, wang2024hpt}, but they do so without a clear picture of what is being lost. A more fundamental question precedes both strategies: \emph{what constitutes a good action representation, and how can we tell when it is degrading?}

What makes a motor representation support flexible generalization has been studied extensively in cognitive neuroscience.
The mirror neuron theory~\citep{dipellegrino1992understanding, gallese1996action, rizzolatti2004mirror} reveals that in primates, the same neurons fire when observing a goal-directed action and when executing it~\citep{fadiga1995motor, keysers2009expanding}.
Crucially, these neurons are organized around \emph{intentions}: they respond differently to identical motor acts embedded in different goals, forming an intention-level ``motor vocabulary''~\citep{fogassi2005parietal, rizzolatti2010parieto}.
This implies that biological motor representations encode not only movement patterns but also \emph{what task the agent intends to accomplish}, and it is this intention-level organization that supports flexible generalization to novel situations.

We draw an analogy to robot learning.
A VLA's action representations do not merely encode motor commands; they also carry the model's understanding of the current task, \eg, whether the robot is picking up a spoon, stacking a dish, or closing a cabinet.
If this understanding remains well-organized, the model can transfer knowledge across similar tasks; if it degrades, the model may fall back on superficial cues and lose generalization.
Under the mirror neuron analogy, pretrained vision-language encoders can be viewed as a concrete instance of the ``observation side'': trained on large-scale multimodal data, they compress diverse visual and linguistic experience into semantic manifolds in which task-relevant concepts are well organized.
The Platonic Representation Hypothesis~\citep{huh2024platonic} suggests that such manifolds reflect a convergent semantic organization rather than an idiosyncratic encoding.
These manifolds provide a reference for examining the intention-level organization of VLA action representations on the ``execution side.''

We design a probing framework (\S\ref{sec:diagnostic}) to test this idea. Tracking the action--semantic alignment across $\pi_0$'s full fine-tuning trajectory, we find that action-only fine-tuning progressively erodes the intention-level structure inherited from pretraining, and that the surviving structure's quality is associated with both out-of-distribution generalization and per-trajectory rollout success.

Motivated by these findings, we propose a training-time method (\S\ref{sec:method}) that contrastively anchors VLA mid-layer action representations to a well-structured semantic manifold (Fig.~\ref{fig:teaser}b).
Because semantic alignment rewards invariance to surface variation while action prediction requires sensitivity to physical detail, we decompose each representation into a \emph{shared} channel aligned to the semantic manifold and a \emph{private} channel that preserves execution-specific detail.
All auxiliary modules are discarded at inference, leaving the deployed model identical to the action-only baseline.

Our contributions are as follows:
\begin{itemize}
\item We propose the semantic structure of action representations as a measurable proxy for representation quality, and show via systematic probing that action-only fine-tuning erodes this structure while its quality is associated with task success and out-of-distribution generalization.
\item We propose a method that restores this structure during training without adding inference cost, combining contrastive alignment to a semantic manifold with shared/private decomposition.
\item We validate on architecturally distinct VLA backbones across simulation and real-world benchmarks, showing up to $+18.7\%$ on real-world in-distribution tasks and $+21.5\%$ on OOD generalization.
\end{itemize}

\section{Diagnostic: Action--Instruction Alignment Erosion}
\label{sec:diagnostic}

\subsection{Probing Setup}
\label{sec:diag:setup}

\begin{wrapfigure}[18]{r}{0.6\textwidth}
\centering
\includegraphics[width=0.6\textwidth]{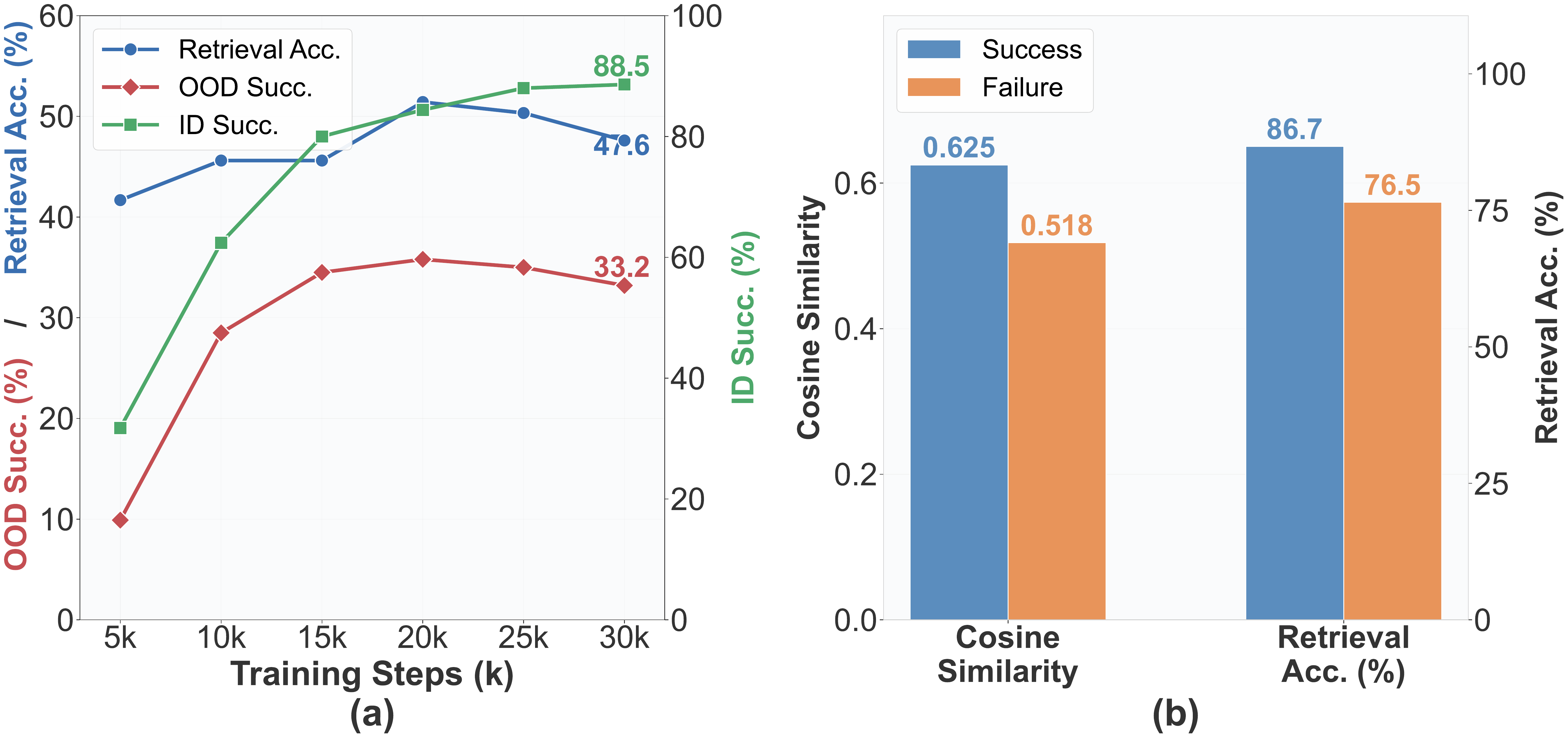}
\caption{\textbf{Alignment probing results on LIBERO~\citep{liu2023libero}.}
\textbf{(a)}~Action--instruction alignment (retrieval accuracy), ID success, and OOD success tracked over the $\pi_0$ fine-tuning trajectory. OOD success follows a similar trend to alignment, while ID success rises monotonically.
\textbf{(b)}~Per-trajectory cosine similarity and retrieval accuracy at step 30k, separated by rollout outcome; both metrics are systematically higher for successful rollouts.}
\label{fig:diag}
\end{wrapfigure}

Action-only fine-tuning may erode the semantic structure inherited from pretraining. To track whether this occurs, we measure the alignment between action representations and instruction embeddings from a separately pretrained encoder across the full fine-tuning trajectory of $\pi_0$~\citep{black2024pi0} on LIBERO~\citep{liu2023libero}, probing at checkpoints from the pretrained VLA through step 30k, and examining whether the trend is associated with downstream performance.

Prior work~\citep{gao2025dontblind,reuss2025flower} shows that the instruction-relevant semantic structure
peaks around the early-to-mid layers. We extract the layer-$k{=}10$ hidden states
$\hraw \in \mathbb{R}^{d}$, out of $N{=}18$ backbone layers, and mean-pool over the
action-token positions into $\hpool$. Following the Platonic Representation Hypothesis~\citep{huh2024platonic}, we use a separately pretrained encoder, the text encoder of Qwen3-VL-Embedding~\citep{li2026qwen3} ($\etext=\gprobe(\instr)$), as a stable semantic reference, since the VLA's own language pathway also degrades during fine-tuning.
To quantify agreement we adopt the alignment-probing protocol
of~\citet{zhang2025assessing}: freeze both sides and train lightweight projection heads
$\probeA, \probeT$ ($\zproj_a=\probeA(\hpool)$, $\zproj_t=\probeT(\etext)$) into a
shared $\dproj$-dim space on task-disjoint LIBERO pairs, using 8 tasks for training and the remaining 2 tasks for evaluation within each of the 4 suites, with
the bidirectional InfoNCE objective~\citep{oord2018infonce},
\begin{equation}
\label{eq:diag-infonce}
\Lprobe = -\frac{1}{2\bsz}\sum_{n=1}^{\bsz}\!\left[\,
\log\!\frac{\exp\!\big(\cossim(\zproj_a^{(n)},\zproj_t^{(n)})/\temp\big)}
         {\sum_{m=1}^{\bsz}\exp\!\big(\cossim(\zproj_a^{(n)},\zproj_t^{(m)})/\temp\big)}
+
\log\!\frac{\exp\!\big(\cossim(\zproj_t^{(n)},\zproj_a^{(n)})/\temp\big)}
         {\sum_{m=1}^{\bsz}\exp\!\big(\cossim(\zproj_t^{(n)},\zproj_a^{(m)})/\temp\big)}
\,\right],
\end{equation}
with $\cossim$ cosine similarity, $\bsz$ the batch size, and $\temp$ a learnable temperature. We report the
bidirectional Recall@1 under the trained heads as the action--instruction alignment at
layer~$k$.

\subsection{Alignment Dynamics over Fine-Tuning}
\label{sec:diag:erosion}

We probe seven checkpoints at 5k-step intervals from the pretrained backbone to
step 30k, measuring the layer-$k$ alignment at each. For the six checkpoints with trained policies, steps 5k--30k, we additionally measure in-distribution success on
LIBERO~\citep{liu2023libero} and out-of-distribution success on
LIBERO-Pro~\citep{zhou2025liberopro}. $\probeA,\probeT$ are retrained at every checkpoint.

Relative to the pretrained backbone, the alignment drops sharply early in fine-tuning,
recovers only partially, declines again late in training, and never returns to its
pretrained level of $62.74\%$ (Fig.~\ref{fig:diag}a, blue); the intention-level semantic structure
inherited from pretraining is progressively eroding. In-distribution success, by
contrast, rises almost monotonically (Fig.~\ref{fig:diag}a, green). This divergence highlights a limitation of in-distribution evaluation:
LIBERO's training and evaluation distributions are nearly identical, so
in-distribution success can be attained by fitting distribution-specific shortcuts. Out-of-distribution success on
LIBERO-Pro tells a different story: it follows a similar rise-then-fall trajectory to
the alignment (Fig.~\ref{fig:diag}a, red; Spearman $\rho{=}0.943$). In summary, action-only fine-tuning progressively erodes the intention-level semantic structure inherited from pretraining, and the surviving structure's quality is associated with out-of-distribution generalization.

\subsection{Per-Trajectory Alignment and Rollout Outcome}
\label{sec:diag:crosssec}

Beyond the trends observed during fine-tuning, we ask whether alignment is also associated with individual rollout outcomes. Using the step-30k policy on LIBERO, we compute per-trajectory cosine similarity between action features and the instruction embedding.
Fig.~\ref{fig:diag}b shows that both cosine similarity and retrieval accuracy are systematically higher for successful than for failed rollouts. Together with the longitudinal trend above, these results motivate an explicit alignment objective: if semantic structure both tracks generalization and reflects per-episode success, actively preserving it during fine-tuning may yield better policies.

\section{Method}
\label{sec:method}

\subsection{Problem Formulation}
\label{sec:method:prelim}

A pretrained VLA $\vla$ maps a visual observation $\obs_t$ and a natural-language instruction $\instr$ to an action $\act_t$. Standard post-training on demonstrations $\Dset=\{(\obs_t,\instr,\act_t)\}$ minimizes an action loss
\begin{equation}
\label{eq:lact}
\Lact = \mathbb{E}_{\Dset}\!\Big[\,\ploss\!\big(\vla(\obs_t,\instr),\;\act_t\big)\Big],
\end{equation}
where $\ploss$ is the backbone's native per-sample action cost, i.e., cross-entropy over discretized action tokens for autoregressive VLAs or a flow-matching regression cost for continuous VLAs~\citep{black2024pi0}.
As \S\ref{sec:diagnostic} shows, this objective provides no mechanism to preserve the inherited semantic structure. We therefore add an alignment objective whose gradient is routed through an auxiliary branch, aiming to reduce interference with action prediction.

We extract the $\nact$ action-token hidden states $\{\hraw_i\}_{i=1}^{\nact}$ at backbone layer~$\layer$, each $\hraw_i \in \mathbb{R}^{d}$. The alignment target is the instruction embedding $\etext = \genc(\instr)$ from the text encoder of a frozen vision-language model EgoHOD~\citep{pei2025egohod}, whose output space serves as a concrete proxy for the semantic manifold introduced in \S\ref{sec:intro}~\citep{huh2024platonic}. We denote this encoder as $\genc$; \S\ref{sec:ablation} ablates the encoder choice.

\subsection{Anchoring Action Features to the Semantic Manifold}
\label{sec:method:align}

\begin{figure}[t]
\centering
\includegraphics[width=0.95\textwidth]{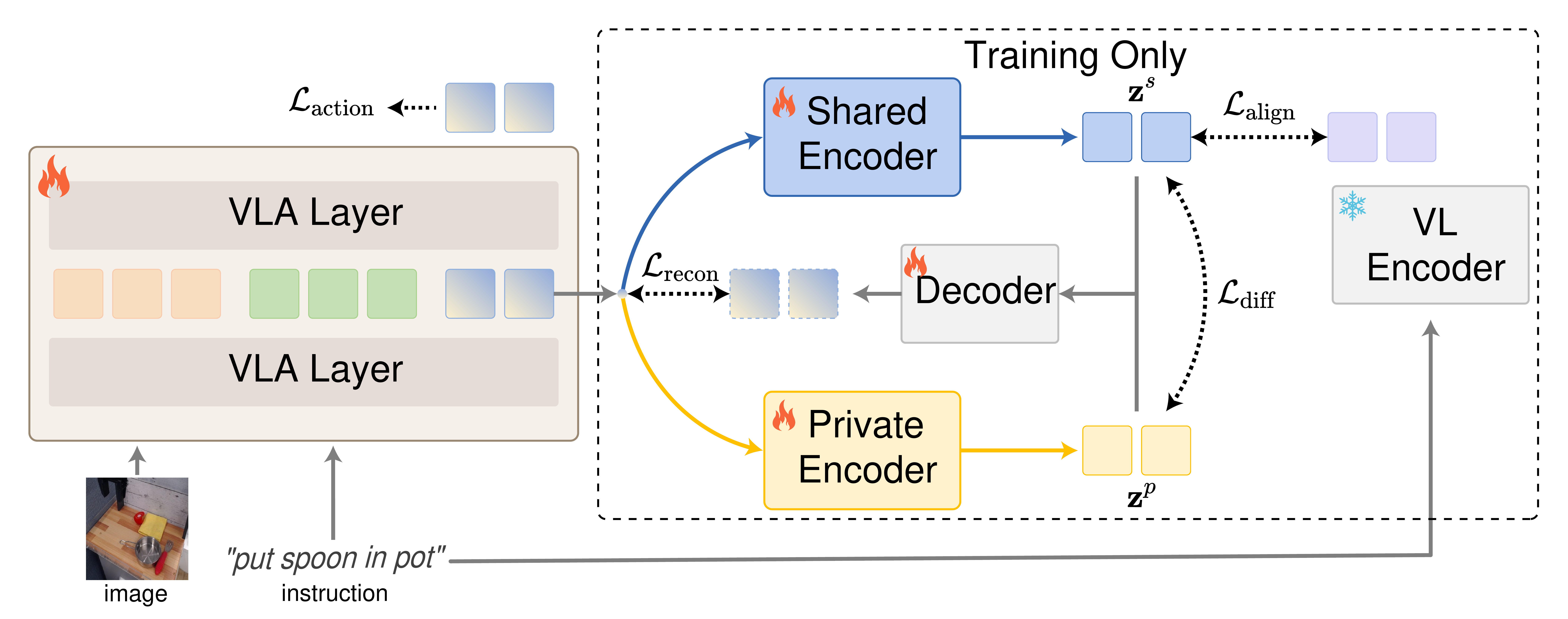}
\caption{\textbf{Method overview.} A pretrained VLA maps visual observations and language instructions to actions. At a mid-network layer, action-token representations are attention-pooled, and the feature is decomposed into a shared component that encodes intention-level semantics and a private component that preserves execution-specific detail. The shared component is contrastively aligned to the instruction embedding from a frozen encoder. All modules inside the dashed box are discarded at inference, leaving the deployed model unchanged.}
\label{fig:method}
\end{figure}

We aggregate the action-token hidden states into a sentence-level representation via cross-attention with a learnable query $\qpool \in \mathbb{R}^{d}$:
\begin{equation}
\label{eq:attn-pool}
\alpha_i = \frac{\exp(\qpool^{\!\top} \hraw_i / \sqrt{d})}
               {\sum_{m=1}^{\nact}\exp(\qpool^{\!\top} \hraw_m / \sqrt{d})},
\qquad
\hpool = \sum_{i=1}^{\nact} \alpha_i\, \hraw_i.
\end{equation}
Two-layer MLP heads $\probeA, \probeT$ project $\hpool$ and $\etext$ into a shared $\dproj$-dimensional space: $\zproj_a=\probeA(\hpool)$, $\zproj_t=\probeT(\etext)$.
We use task IDs to identify semantic clusters of language instructions and construct a positive-pair mask. For each action representation, instructions in the same cluster are treated as positives, while those in other clusters within the batch serve as negatives. We apply a multi-positive pairwise contrastive loss~\citep{zhai2023sigmoid}:
\begin{equation}
\label{eq:lalign}
\Lalign = -\frac{1}{\bsz^{2}}\sum_{n=1}^{\bsz}\sum_{m=1}^{\bsz}
\log\sigma\!\Big(y_{nm}\big(\cossim(\zproj_a^{(n)},\zproj_t^{(m)})/\temp - b\big)\Big),
\end{equation}
where $\sigma$ is the logistic function, $y_{nm}{=}1$ for positive pairs and $-1$ otherwise. The alignment gradient flows through $\probeA$ into the backbone at layer~$\layer$, actively shaping the representation. The pooled representation aggregates both intention-level and execution-specific information; the next section refines this by decomposing $\hpool$ so that only the intention-level component enters $\Lalign$.

\subsection{Shared--Private Decomposition}
\label{sec:method:decompose}

Semantic alignment rewards \emph{invariance} to surface variation, while action prediction requires \emph{sensitivity} to physical detail, opposing requirements that loss weighting alone cannot resolve. Inspired by the separation of goal-level and effector-specific coding in the mirror-neuron circuit~\citep{gallese1996action,rizzolatti2004mirror}, we decompose the attention-pooled feature into a \emph{shared} channel for intention-level semantics and a \emph{private} channel that captures the execution-specific residual, including embodiment geometry, timing, and motor detail, needed for action prediction but orthogonal to task semantics. The decomposition follows Domain Separation Networks~\citep{bousmalis2016dsn}.

Concretely, two-layer MLP encoders decompose the feature $\hpool$ into a $\dsub$-dimensional shared representation $\zs$ and a $\dsub$-dimensional private representation $\zp$; a decoder produces the reconstruction ${\hrec}$ additively:
\begin{equation}
\label{eq:dsn-split}
\zs = \Esh(\hpool),\qquad
\zp = \Epr(\hpool),\qquad
{\hrec} = \Dec(\zs + \zp).
\end{equation}
Only the shared component enters the alignment pathway: $\zproj_a = \probeA(\zs)$, so the alignment loss $\Lalign$ receives an intention-level representation free of private-channel detail.
A reconstruction loss ensures the two channels jointly preserve $\hpool$, combining MSE with a squared mean-residual penalty~\citep{bousmalis2016dsn}:
\begin{equation}
\label{eq:lrecon}
\Lrecon =
\tfrac{1}{d}\big\|{\hrec} - \hpool\big\|_{2}^{2}
+ \tfrac{1}{d^{2}}\Big(\sum_{j=1}^{d}\big(h_{\mathrm{rec},j}-\bar{h}_{j}\big)\Big)^{\!2}.
\end{equation}
An angular decorrelation penalty keeps the two channels from re-entangling:
\begin{equation}
\label{eq:ldiff}
\Ldiff = \tfrac{1}{\bsz}\sum_{n=1}^{\bsz}\!\left(\tfrac{\langle\zs_{n},\,\zp_{n}\rangle}{\|\zs_{n}\|\,\|\zp_{n}\|}\right)^{\!2}.
\end{equation}

\subsection{Training Objective}
\label{sec:method:objective}

The full objective combines the action loss with the three alignment terms:
\begin{equation}
\label{eq:ltotal}
\Ltotal = \Lact + \lamA\,\Lalign + \lamR\,\Lrecon + \lamD\,\Ldiff,
\end{equation}
with $\lamR{=}0.001$, $\lamD{=}0.15$. $\lamA$ is backbone-dependent; concrete values are given in Appendix~\ref{app:impl}. All terms are computed over the demonstrations $\Dset$ and require no additional data at training time. At deployment all auxiliary modules are dropped, so the inference graph is identical to the action-only baseline~\citep{yu2025repa,li2025spatialforcing}.

\section{Experiments}
\label{sec:exp}

We design experiments to answer: (RQ1) Does the alignment objective deliver consistent gains across simulation (\S\ref{sec:sim}) and a real bimanual robot (\S\ref{sec:real})? (RQ2) Does the gain transfer across architecturally distinct VLA backbones (\S\ref{sec:sim})? (RQ3) Which design choices drive the gain (\S\ref{sec:ablation})?

\subsection{Simulation Experiments}
\label{sec:sim}

\paragraph{Setup}
We evaluate on two simulation benchmarks that together cover small-data in-domain capability and large-scale cross-domain transfer. LIBERO~\citep{liu2023libero} contains four task suites with 50 demonstrations per task and tests near-distribution performance. SimplerEnv~\citep{li2024simpler} fine-tunes backbones on BridgeData V2~\citep{walke2023bridgedata}, a large-scale, multi-source real-robot dataset, and evaluates WidowX manipulation tasks under the Visual Matching protocol.

Our main experiments use three architecturally distinct backbones: $\pi_0$~\citep{black2024pi0}, a flow-matching VLA that we fully fine-tune; SpatialVLA~\citep{qu2025spatialvla}, an autoregressive VLA that we fine-tune with LoRA; and GR00T N1.6~\citep{bjorck2025gr00t}, a dual-system VLA with a diffusion-transformer action head. We follow the official recipe of each backbone for the action objective and add only our alignment objective. The frozen alignment target is the text encoder of EgoHOD~\citep{pei2025egohod}, applied at layer $\layer{=}10$ for $\pi_0$, SpatialVLA, and GR00T N1.6. All alignment modules are discarded at inference, so the deployed policy is identical to the action-only baseline. Detailed hyperparameters are in Appendix~\ref{app:impl}.

\begin{table}[t]
\centering
\small
\begin{subtable}[b]{0.44\textwidth}
\centering
\renewcommand{\arraystretch}{1.4448}%
\resizebox{\linewidth}{!}{%
\begin{tabular}{lccccc}
\toprule
Method & Spatial & Object & Goal & Long & Avg \\
\midrule
OpenVLA~\citep{kim2024openvla}             & 84.7 & 88.4 & 79.2 & 53.7 & 76.5 \\
Octo~\citep{ghosh2024octo}                 & 78.9 & 85.7 & 84.6 & 51.1 & 75.1 \\
$\pi_0$-FAST~\citep{pertsch2025pi0fast}    & 96.4 & 96.8 & 88.6 & 60.2 & 85.5 \\
SmolVLA~\citep{shukor2025smolvla}          & 90.0 & 96.0 & 92.0 & 71.0 & 87.3 \\
\midrule
$\pi_0$~\citep{black2024pi0}              & 94.0 & 96.2 & 90.8 & 73.0 & 88.5 \\
$\pi_0$ + Ours                             & \textbf{95.7} & \textbf{97.7} & \textbf{91.7} & \textbf{78.3} & \textbf{90.8}\up{2.3} \\
\bottomrule
\end{tabular}%
}
\caption{LIBERO~\citep{liu2023libero}.}
\label{tab:libero}
\end{subtable}
\hfill
\begin{subtable}[b]{0.49\textwidth}
\centering
\resizebox{\linewidth}{!}{%
\begin{tabular}{lccccc}
\toprule
Method & spoon & carrot & stack & eggplant & Avg \\
\midrule
Octo~\citep{ghosh2024octo}                 & 47.2 & 9.7  & 4.2  & 56.9  & 29.5 \\
$\pi_0$-FAST~\citep{pertsch2025pi0fast}    & 29.1 & 21.9 & 10.8 & 66.6  & 32.1 \\
\midrule
$\pi_0$~\citep{black2024pi0}               & 36.1 & 33.3 & 23.6 & 51.4  & 36.1 \\
$\pi_0$ + Ours                             & \textbf{48.6} & \textbf{37.5} & \textbf{31.9} & \textbf{56.9}  & \textbf{43.8}\up{7.7} \\
\midrule
SpatialVLA~\citep{qu2025spatialvla}        & 23.6 & 29.2 & 23.6 & 95.8 & 43.1 \\
SpatialVLA + Ours                          & \textbf{36.1} & \textbf{40.3} & \textbf{27.8} & \textbf{97.2} & \textbf{50.3}\up{7.2} \\
\midrule
GR00T N1.6~\citep{bjorck2025gr00t}         & 84.7 & 50.0 & 5.6 & \textbf{98.6} & 59.7 \\
GR00T N1.6 + Ours                          & \textbf{87.5} & \textbf{51.4} & \textbf{23.6} & \textbf{98.6} & \textbf{65.3}\up{5.6} \\
\bottomrule
\end{tabular}%
}
\caption{SimplerEnv~\citep{li2024simpler}.}
\label{tab:simpler}
\end{subtable}
\caption{Simulation success rates (\%) on LIBERO~\citep{liu2023libero} and SimplerEnv~\citep{li2024simpler}. Semantic anchoring improves performance across benchmarks and VLA backbones. \textbf{Bold}: improvement over the respective baseline.}
\label{tab:sim_combined}
\end{table}

\paragraph{Simulation Results}
Table~\ref{tab:sim_combined} summarizes the results. On LIBERO, our alignment objective improves accuracy on every suite despite only 50 demonstrations per task, indicating that the alignment objective does not impair action prediction and provides complementary gains even in near-distribution settings. The t-SNE visualizations on LIBERO-Goal (Fig.~\ref{fig:teaser}a,c) confirm this.
On SimplerEnv, which evaluates under a real-to-sim domain shift, gains are larger across all three backbones, likely because BridgeData V2's richer task diversity gives the alignment more semantic material to anchor onto. LIBERO results for $\pi_0$, and SimplerEnv results for all three backbones, are averaged over three training seeds; standard deviations are reported in Appendix~\ref{app:backbone-seeds}.
We further evaluate on LIBERO-Pro~\citep{zhou2025liberopro}, an extended perturbation benchmark built on LIBERO, and find that our method consistently improves $\pi_0$ on every perturbation axis; full results are in Appendix~\ref{app:libero-pro}.

\newpage
\subsection{Real-Robot Experiments}
\label{sec:real}

\begin{figure}[t]
\centering
\includegraphics[width=\textwidth]{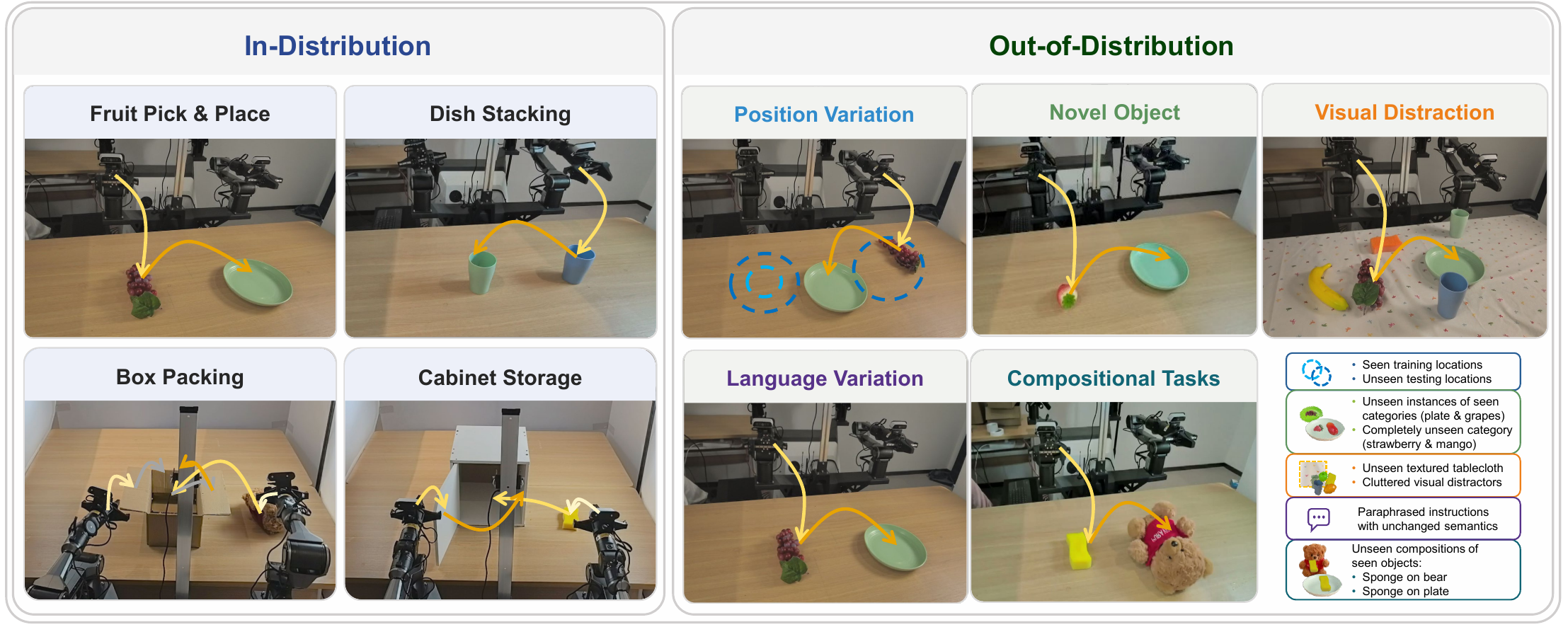}
\caption{Real-robot experimental setup. \textbf{Left:} four in-distribution task families (two single-stage, two with articulated closure). \textbf{Right:} five out-of-distribution axes built on the pick-and-place family.}
\label{fig:real_setup}
\end{figure}

\paragraph{Setup}
The platform is a bimanual AgileX Cobot Magic V2 with two 6-DoF arms, parallel grippers, and three RGB cameras, including one front-view and two wrist-mounted. We design four task families with two variants each (Fig.~\ref{fig:real_setup}, left): \emph{fruit pick-and-place} and \emph{dish stacking} are single-stage tasks, while \emph{box packing} and \emph{cabinet storage} each chain object placement with an articulated closure by closing the box flaps or the cabinet door, requiring longer-horizon coordination.

We evaluate generalization along five axes on the pick-and-place family (Fig.~\ref{fig:real_setup}, right): \emph{Language}, which uses paraphrased instructions; \emph{Position}, with unseen layouts; \emph{Object}, introducing novel objects; \emph{Visual}, adding new backgrounds and distractors; and \emph{Task}, which tests unseen source--target combinations of objects seen during training.

We collect 200 teleoperated trajectories per task and fully fine-tune a single multi-task $\pi_0$ policy; our method shares this protocol and adds the alignment objective at the same mid-network layer as in simulation. Baselines include the same multi-task $\pi_0$ trained under the same protocol, as well as per-task ACT~\citep{zhao2023act} and Diffusion Policy~\citep{chi2023diffusion}, trained individually because their multi-task variants did not converge in our setup. Because ACT and Diffusion Policy are trained per task without language conditioning, the Language and Task axes are not applicable in Table~\ref{tab:real}. Each task is evaluated for 20 trials. Full hyperparameters are provided in Appendix~\ref{app:platform}.

\paragraph{Results}
Table~\ref{tab:real} summarizes both in-distribution and out-of-distribution results. Our method consistently improves $\pi_0$ across all four task families. For out-of-distribution generalization, improvements are again consistent across all five axes, with the largest gains on Position and Task, the two most challenging axes that demand spatial and compositional reasoning beyond memorized demonstrations. On the Task axis, which requires executing unseen source--target combinations of objects seen during training, $\pi_0$ frequently fails to interpret the novel composition, stalling or grasping irrelevant objects, whereas our method correctly identifies the target and selects the appropriate arm; see Appendix~\ref{app:qualitative} for qualitative comparisons.

\begin{table}[t]
\centering
\small
\setlength{\tabcolsep}{2.2pt}
\begin{tabular}{l ccccc @{\hspace{7pt}}|@{\hspace{5pt}} cccccc c}
\toprule
& \multicolumn{5}{c@{\hspace{7pt}}|@{\hspace{5pt}}}{In-Distribution}
& \multicolumn{7}{c}{Out-of-Distribution} \\
Method
& \scriptsize\shortstack{Fruit\\Pick\&Place}
& \scriptsize\shortstack{Dish\\Stacking}
& \scriptsize\shortstack{Box\\Packing}
& \scriptsize\shortstack{Cabinet\\Storage}
& Avg
& Orig.
& Lang.
& Pos.
& Obj.
& Vis.
& Task
& Avg \\
\midrule
ACT
& 82.5 & 45.0 & 25.0 & 47.5 & 50.0
& 82.5 & --   & 7.5  & 40.0 & 45.0 & --   & -- \\
DP
& 80.0 & 47.5 & 32.5 & 42.5 & 50.6
& 80.0 & --   & 12.5 & 32.5 & 35.0 & --   & -- \\
$\pi_0$
& 75.0 & 42.5 & 37.5 & 50.0 & 51.3
& 75.0 & 75.0 & 32.5 & 52.5 & 57.5 & 30.0 & 49.5 \\
$\pi_0$ + Ours
& \textbf{92.5} & \textbf{67.5} & \textbf{52.5} & \textbf{67.5} & \textbf{70.0}
& \textbf{92.5} & \textbf{85.0} & \textbf{57.5} & \textbf{77.5} & \textbf{75.0} & \textbf{60.0} & \textbf{71.0} \\
\bottomrule
\end{tabular}
\vspace{4pt}
\caption{Real-robot success rate (\%, 20 trials each). \textbf{Left:} in-distribution across four task families. \textbf{Right:} out-of-distribution generalization evaluated on the pick-and-place family.}
\label{tab:real}
\end{table}

\subsection{Ablation and Analysis}
\label{sec:ablation}

We ablate one design choice at a time on SpatialVLA evaluated in SimplerEnv, fixing the training protocol of Table~\ref{tab:simpler}.

\begin{figure}[ht]
\centering
\includegraphics[width=\textwidth]{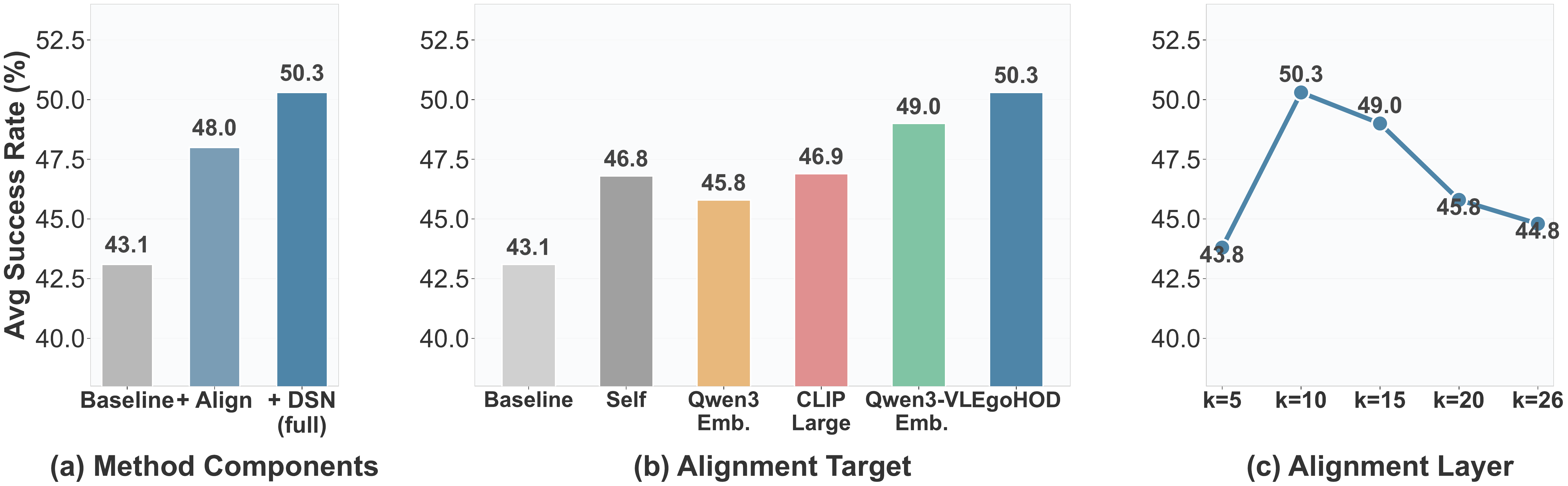}
\caption{\textbf{Ablation study} with SpatialVLA~\citep{qu2025spatialvla} on SimplerEnv~\citep{li2024simpler}. \textbf{(a)}~Method components: \emph{+\,Align}~applies contrastive alignment without decomposition; \emph{+\,DSN}~adds shared/private decomposition. \textbf{(b)}~Alignment target: \emph{Self} contrastively aligns same-task action features without an external encoder; the remaining bars use different pretrained encoders. \textbf{(c)}~Performance by alignment layer $k$.}
\label{fig:ablation}
\end{figure}

\paragraph{Method Components}
We incrementally add each module to isolate its contribution: contrastive alignment alone vs.\ the full method with shared/private decomposition. As shown in Fig.~\ref{fig:ablation}a, each component contributes incrementally. Applying contrastive alignment alone, \emph{+\,Align}, already lifts the baseline, confirming that anchoring action features to a semantic reference is beneficial even without explicit channel separation. Adding the shared/private decomposition, \emph{+\,DSN}, further improves the average, indicating that decomposing intention-level semantics from execution-specific detail provides a structural prior that strengthens alignment beyond what the contrastive loss alone can achieve.

\paragraph{Alignment Target}
We vary the frozen encoder $\genc$ to examine the impact of the alignment target (Fig.~\ref{fig:ablation}b). The \emph{Self} variant contrastively clusters same-task action features without an external encoder; it improves over the baseline but falls well below EgoHOD, indicating that anchoring to a semantically rich external space provides substantially stronger guidance than task-identity clustering alone. Among external encoders, visual grounding is critical: the language-only Qwen3-Embedding underperforms every visually grounded alternative. Within the visually grounded group, CLIP-Large, trained on static image-text pairs, underperforms both Qwen3-VL-Embedding and EgoHOD, which incorporate temporal video structure; since EgoHOD shares the same text encoder as CLIP-Large, the gap isolates the contribution of visual dynamics. Among the two video-grounded encoders, EgoHOD outperforms the larger Qwen3-VL-Embedding, confirming that domain-specific pretraining on manipulation videos matters more than model scale.

\paragraph{Alignment Layer}
We sweep the backbone layer $k$ at which the alignment objective is attached to identify where semantic structure is richest. Fig.~\ref{fig:ablation}c shows results across the full depth. Mid-network layers yield the largest gains, with performance peaking at $k{=}10$ and remaining strong at $k{=}15$; both early and late layers diminish the benefit. This is consistent with prior findings that mid-layer representations retain the richest semantic structure, while later layers increasingly specialize for action prediction~\citep{kim2025rscl, gao2025dontblind}. We set $k{=}10$ for all three backbones.

\newpage
\section{Related Work}
\label{sec:related}

\paragraph{Vision-Language-Action Models} VLA models fine-tune pretrained VLMs on robot demonstrations, inheriting semantic representations that enable generalization~\citep{zitkovich2023rt,kim2024openvla,black2024pi0}. The paradigm has been extended with generalist diffusion heads~\citep{ghosh2024octo}, flow-matching decoders~\citep{black2024pi0}, 3D spatial conditioning~\citep{qu2025spatialvla}, and cross-embodiment trunks~\citep{wang2024hpt}. A recurring observation is that action supervision on narrow data distorts pretrained features~\citep{kumar2022finetuning,aghajanyan2021r3f}, inducing shortcut learning~\citep{geirhos2020shortcut,li2025shortcutrobot,gao2025dontblind}. Existing remedies scale data~\citep{intelligence2025pi05,bjorck2025gr00t}, redesign architectures~\citep{qu2025spatialvla,ghosh2024octo}, or add auxiliary objectives~\citep{kim2025rscl,gao2025dontblind}. Distinct from these approaches, our method does not scale data or modify the model architecture, but instead establishes alignment with a semantic manifold as a measurable criterion for representation quality and repairs the eroded structure through a training-time-only objective.

\paragraph{Representation Alignment} Cross-modal contrastive alignment, exemplified by CLIP~\citep{radford2021clip}, has been widely used to learn transferable representations, with the Platonic Representation Hypothesis~\citep{huh2024platonic} providing theoretical motivation. REPA~\citep{yu2025repa} aligns diffusion-transformer hidden states to a frozen encoder at training time only; our method inherits this design but adapts it to VLAs with a manipulation-centric encoder and a shared/private decomposition designed to mitigate gradient conflict between semantic alignment and action prediction. In the embodied domain, contrastive pretraining on egocentric video has yielded reusable representations~\citep{nair2022r3m,ma2023vip,karamcheti2023voltron}, and EgoHOD~\citep{pei2025egohod} captures fine-grained hand--object dynamics that we leverage as a frozen semantic reference. Concurrent work preserves VLA representations via visual-pathway regularization~\citep{gao2025dontblind}, self-feature contrastive objectives~\citep{kim2025rscl}, or spatial alignment~\citep{li2025spatialforcing}; our approach differs in targeting action-token representations and anchoring to an external semantic manifold. Zhu et al.~\citep{zhu2025mirror} share our mirror-neuron motivation in an embodied setting; our work focuses on the VLA fine-tuning regime with diagnostic evidence linking alignment quality to performance.

\section{Conclusion, Limitations and Future Directions}
\label{sec:conclusion}

We show that action-only fine-tuning of VLAs erodes the intention-level semantic structure inherited from pretraining, and that this structure's quality is associated with generalization. Guided by this finding and the mirror neuron theory, we propose a method that anchors action representations to a semantic manifold via shared/private decomposition during training, with all auxiliary modules discarded at inference. Experiments on three architecturally distinct VLA backbones across simulation and real-world settings validate consistent improvements in both in-distribution and out-of-distribution settings. Despite these promising results, several limitations remain. First, performance is upper-bounded by the quality of the frozen alignment target; a stronger manipulation-centric encoder would likely yield further gains. Second, the shared/private decomposition constrains subspace relations without identifiability guarantees. Third, the task-level alignment loss ignores graded cross-task similarity and token-level semantics. Finally, we apply the objective only during downstream fine-tuning of pretrained VLAs on task-specific demonstrations. Future work will explore extending semantic alignment to the VLA pretraining stage and scaling to broader data sources toward more generalizable robotic manipulation.

\clearpage
\acknowledgments{This work was supported by the National Key Research and Development Program of China (No. 2026ZD0222204), the National Natural Science Foundation of China (No. 6247070125), and the Zhejiang Provincial Natural Science Foundation of China under Grant No. LQN26F030037. The authors also thank the High Performance Computing Center at Eastern Institute of Technology, Ningbo, and the Ningbo Institute of Digital Twin for providing computational resources.}

\bibliography{references}

\clearpage
\appendix

\section{Implementation Details}
\label{app:impl}

\subsection{Backbone Training Hyperparameters}
\label{app:backbone}

We follow the official training recipe of each backbone and add only our alignment objective. Table~\ref{tab:backbone_hparams} lists backbone-specific hyperparameters for all five experimental settings.

$\pi_0$~\citep{black2024pi0} uses PaliGemma-2B as the vision-language backbone and Gemma-300M as the action expert. All parameters are fully fine-tuned. On LIBERO, we train for 30k steps with a peak learning rate of $2.5{\times}10^{-5}$ and a batch size of 32. On BridgeData~V2, we train for 40k steps with a peak learning rate of $5{\times}10^{-5}$ and a batch size of 128. For the real robot, we train a single multi-task policy on all eight task variants for 30k steps with a peak learning rate of $5{\times}10^{-5}$ and a batch size of 64.

SpatialVLA~\citep{qu2025spatialvla} is a 4B-parameter autoregressive VLA. On BridgeData~V2, we fine-tune it with LoRA of rank~32 and $\alpha{=}32$ applied to all linear layers for 10k steps, with a peak learning rate of $10^{-4}$. We use DeepSpeed ZeRO Stage-1 for memory efficiency.

GR00T N1.6~\citep{bjorck2025gr00t} is a 3B-parameter dual-system VLA with a diffusion-transformer action head. On BridgeData~V2, we follow the official fine-tuning recipe: the vision-language backbone is kept frozen while the projector and the diffusion-transformer action head are fine-tuned for 15k steps with a peak learning rate of $10^{-4}$, a warmup ratio of 0.05, and a global batch size of 1024.

All experiments use AdamW with gradient clipping at norm~1.0 and bfloat16 mixed precision.

\begin{center}
\small
\setlength{\tabcolsep}{4pt}
\resizebox{\linewidth}{!}{%
\begin{tabular}{lccccc}
\toprule
& $\pi_0$ (LIBERO) & $\pi_0$ (BridgeV2) & SpatialVLA (BridgeV2) & GR00T N1.6 (BridgeV2) & $\pi_0$ (Real) \\
\midrule
Fine-tuning     & Full          & Full          & LoRA ($r{=}32$) & Projector + DiT head & Full \\
Backbone LR     & $2.5\!{\times}\!10^{-5}$ & $5\!{\times}\!10^{-5}$ & $10^{-4}$ & $10^{-4}$ & $5\!{\times}\!10^{-5}$ \\
Training steps  & 30k           & 40k           & 10k             & 15k & 30k \\
Batch size      & 32            & 128           & 192             & 1024 & 64 \\
$\beta_1, \beta_2$ & 0.9, 0.95 & 0.9, 0.95    & 0.9, 0.999      & 0.9, 0.999 & 0.9, 0.95 \\
Weight decay    & $10^{-10}$    & $10^{-10}$    & 0               & $10^{-5}$ & $10^{-10}$ \\
Action horizon  & 50            & 5             & 4               & 8 & 50 \\
VLM frozen      & --            & --            & LoRA only       & Frozen & -- \\
GPUs            & 2             & 2             & 2               & 4 & 2 \\
Subspace dim.\ $\dsub$ & 512     & 512           & 1024            & 512 & 512 \\
\bottomrule
\end{tabular}%
}
\captionof{table}{Backbone training hyperparameters.}
\label{tab:backbone_hparams}
\end{center}

\subsection{Alignment Module}
\label{app:align-module}

At training time, the alignment module contains an attention-pooling block, shared/private encoders, a reconstruction decoder, and lightweight projection heads. Given the backbone features $\hraw_i \in \mathbb{R}^{d}$, a learnable query $\qpool \in \mathbb{R}^{d}$ in the attention-pooling block first aggregates them into a sentence-level representation $\hpool$. The encoders $\Esh$ and $\Epr$ then produce shared and private features $\zs,\zp \in \mathbb{R}^{\dsub}$ from $\hpool$. Both channels use the same subspace dimension $\dsub$, allowing the decoder $\Dec$ to reconstruct the pooled feature from their elementwise sum $\zs + \zp$. The shared action feature $\zs$ and the frozen 768-dimensional EgoHOD-Large text embedding are then projected into the shared contrastive space by lightweight heads $\probeA$ and $\probeT$. The decorrelation term $\Ldiff$ is the mean squared cosine similarity between $\zs$ and $\zp$. All three backbones use layer 10 for alignment, the subspace dimension $\dsub$ listed in Table~\ref{tab:backbone_hparams}, a 512-dimensional projection space, reconstruction weight 0.001, decorrelation weight 0.15, weight decay 0.01, and learnable-query attention with eight heads for pooling. The backbone feature dimension is 1024 for $\pi_0$, 2304 for SpatialVLA, and 1536 for GR00T N1.6. The alignment weight $\lamA$ is 0.1 for $\pi_0$ and 1.0 for SpatialVLA and GR00T N1.6, and the alignment learning rate is $10^{-4}$ for $\pi_0$ and GR00T N1.6 and $5{\times}10^{-5}$ for SpatialVLA. The reconstruction and decorrelation weights are chosen with reference to Domain Separation Networks~\citep{bousmalis2016dsn} and kept identical across backbones. $\lamA$ is set so that the alignment-loss magnitude roughly matches the action loss. In Eq.~\ref{eq:lalign}, the temperature $\temp{=}0.1$ and bias $b{=}1.5$ are shared by all backbones and are fixed rather than learned.

\subsection{Diagnostic Probing Protocol}
\label{app:diag-probing}

The diagnostic probe uses the text encoder of Qwen3-VL-Embedding~\citep{li2026qwen3} as the frozen semantic reference $\gprobe$, deliberately choosing a different model from the EgoHOD alignment target $\genc$ to avoid circularity. We extract layer-10 hidden states and mean-pool over the $\nact$ action-token positions. Two-layer MLP projection heads map the 1024-dimensional action features to a 512-dimensional contrastive space. They are trained with bidirectional InfoNCE on task-disjoint LIBERO pairs, using 8 tasks for training and the remaining 2 tasks for evaluation in each of the four suites. We retrain the heads independently at seven checkpoints: the pretrained backbone and checkpoints from 5k to 30k steps at 5k intervals. Each probe is trained with AdamW for 1k optimization steps using learning rate $10^{-4}$ and weight decay 0.01. We report bidirectional Recall@1 as the alignment metric.

\subsection{Real-Robot Platform}
\label{app:platform}

The platform is a bimanual AgileX Cobot Magic~V2 with two 6-DoF arms and parallel grippers, giving 14 total DoF or 7 per arm including the gripper. Three RGB cameras provide the visual input at $256{\times}256$ resolution: one front-view overhead camera and two wrist-mounted cameras. The action space consists of 14-dimensional joint positions, normalized per-dimension using dataset statistics. We collect 200 teleoperated demonstrations per task variant across four families, giving eight variants and about 1,600 trajectories in total. A single multi-task $\pi_0$ policy is trained on all variants. Each condition is evaluated over 20 trials.

\paragraph{Baseline Training}
ACT~\citep{zhao2023act} and Diffusion Policy~\citep{chi2023diffusion} are trained individually per task, as their multi-task variants did not converge in our bimanual setup. ACT uses a Transformer encoder--decoder with hidden dimension 512, 8 attention heads, 4 encoder layers, 7 decoder layers, a prediction horizon of 50, and dropout 0.1. We train it for 100k steps with batch size 64 and learning rate $10^{-4}$. Diffusion Policy is trained for 30k steps with batch size 16 and learning rate $10^{-4}$. Its action encoder and decoder both use dimension 14. Both baselines use the same 200 demonstrations per task as $\pi_0$.

\subsection{Real-Robot Task Descriptions}
\label{app:real-tasks}

\begin{center}
    \includegraphics[width=\textwidth]{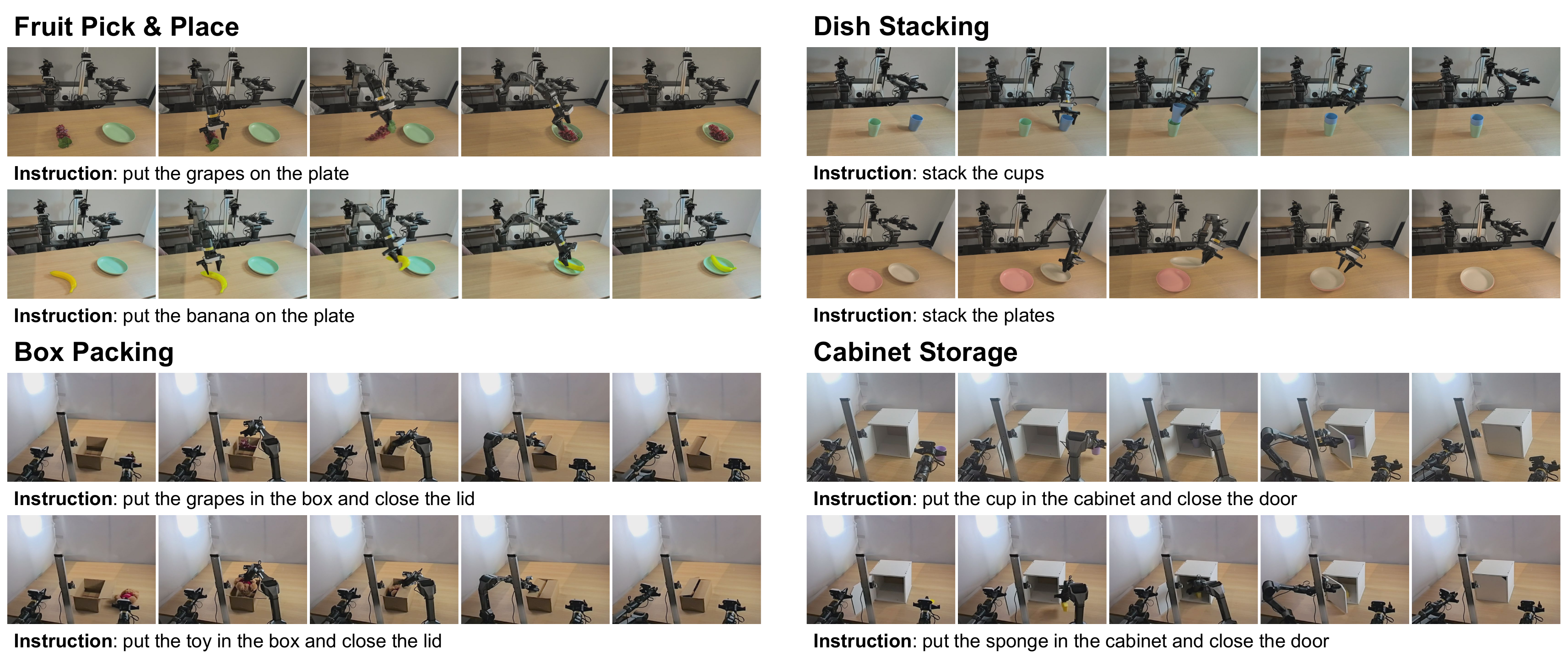}
    \captionof{figure}{Four real-world bimanual task families with two variants each.}
    \label{fig:real_tasks}
\end{center}

Fig.~\ref{fig:real_tasks} illustrates the four task families with two variants each. \emph{Fruit pick-and-place} and \emph{dish stacking} are single-stage tasks, while \emph{box packing} and \emph{cabinet storage} each combine object placement with an articulated closure. Table~\ref{tab:real_tasks} provides detailed task descriptions and evaluation protocols for all eight variants.

\newpage
\begin{center}
\small
\setlength{\tabcolsep}{3pt}
\begin{tabular}{llp{5.2cm}p{4.0cm}}
\toprule
Family & Variant & Task Description & Success Criterion \\
\midrule
\multirow{2}{*}{\shortstack[l]{Fruit\\Pick-and-Place}}
  & Banana & Pick up the banana and place it on the plate. & Fruit rests stably on the plate. \\
  & Grapes  & Pick up the grapes and place it on the plate.  & Fruit rests stably on the plate. \\
\midrule
\multirow{2}{*}{\shortstack[l]{Dish\\Stacking}}
  & Cup   & Stack one cup on top of the other.   & Two cups form a stable upright stack on the table. \\
  & Plate & Stack the plate onto the other plate. & Plate rests stably on the bottom plate. \\
\midrule
\multirow{2}{*}{\shortstack[l]{Box\\Packing}}
  & Grapes & Place the grapes into the box and close both side flaps. & Object inside the box and both side flaps fully closed. \\
  & Bear  & Place the toy bear into the box and close both side flaps. & Object inside the box and both side flaps fully closed. \\
\midrule
\multirow{2}{*}{\shortstack[l]{Cabinet\\Storage}}
  & Cup    & Place the cup in the cabinet and close the door.    & Object inside the cabinet and door closed. \\
  & Sponge & Place the sponge in the cabinet and close the door.  & Object inside the cabinet and door closed. \\
\bottomrule
\end{tabular}
\captionof{table}{Real-world task descriptions and evaluation protocols. Each variant is evaluated over 20 trials with 200 teleoperated demonstrations collected per variant.}
\label{tab:real_tasks}
\end{center}

\subsection{OOD Evaluation Axes}
\label{app:real-ood}

\begin{center}
    \includegraphics[width=\textwidth]{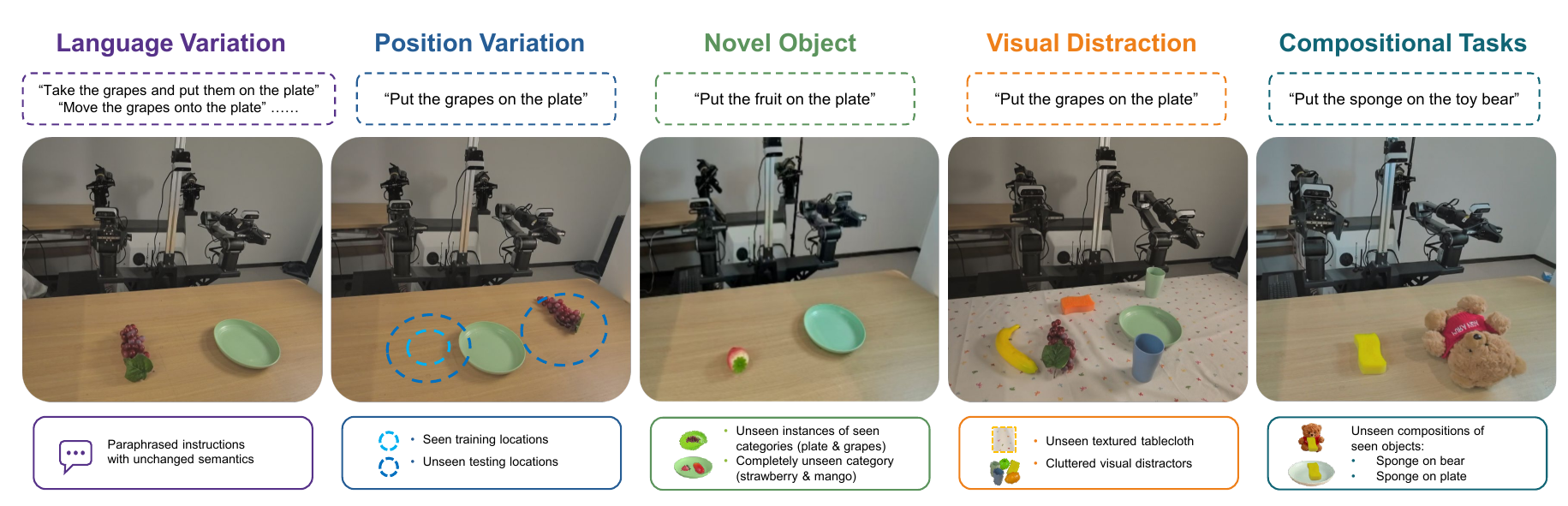}
    \captionof{figure}{Five out-of-distribution generalization axes on the pick-and-place family.}
    \label{fig:real_ood}
\end{center}

All five OOD axes are built on the fruit pick-and-place family (Fig.~\ref{fig:real_ood}). Table~\ref{tab:ood_axes} details the specific perturbations applied in each axis.

\begin{center}
\small
\setlength{\tabcolsep}{3pt}
\renewcommand{\arraystretch}{1.15}
\begin{tabular}{@{}p{1.8cm}p{10cm}@{}}
\toprule
Axis & Description \\
\midrule
Language  & The task instruction is paraphrased into semantically equivalent but lexically different expressions not seen during training. \\
Position  & Objects are placed at positions beyond the spatial range seen during training. \\
Object    & Both in-category and out-of-category perturbations are applied. In-category: the same object type with different attributes (e.g., plates of different shapes, grapes of different sizes). Out-of-category: entirely novel objects not present in any training demonstration (e.g., mango, strawberry). \\
Visual    & The scene background is changed by placing unseen tablecloths, and distractor objects not present during training are added to the workspace. \\
Task      & The policy must execute unseen source--target combinations of objects seen during training. \\
\bottomrule
\end{tabular}
\captionof{table}{Detailed descriptions of the five out-of-distribution evaluation axes.}
\label{tab:ood_axes}
\end{center}

\newpage
\section{Additional Analyses}
\label{app:analyses}

\subsection{LIBERO-Pro Results}
\label{app:libero-pro}

LIBERO-Pro tests perturbation generalization across five axes: language paraphrase, object swap, task composition, environment change, and object replacement. Table~\ref{tab:libero_pro} shows the results.

\begin{center}
\small
\begin{tabular}{lcccccc}
\toprule
Method & lan & swap & task & env & object & Avg \\
\midrule
$\pi_0$        & 61.0 & 1.2 & 6.8  & 44.8 & 66.0 & 36.0 \\
$\pi_0$ + Ours & \textbf{65.0} & \textbf{2.0} & \textbf{15.2} & \textbf{48.2} & \textbf{72.2} & \textbf{40.5} \\
\bottomrule
\end{tabular}
\captionof{table}{LIBERO-Pro success rate (\%). The same trained checkpoint is evaluated without additional training.}
\label{tab:libero_pro}
\end{center}

\subsection{Multi-Seed Evaluation}
\label{app:backbone-seeds}

Each condition uses 20 rollouts per LIBERO task, 24 episodes per SimplerEnv task, or 20 real-robot trials. Table~\ref{tab:multi_seed} reports results over three training seeds as mean $\pm$ standard deviation.

\begin{center}
\small
\begin{tabular}{lcccc}
\toprule
& LIBERO & \multicolumn{3}{c}{SimplerEnv} \\
\cmidrule(lr){2-2}\cmidrule(lr){3-5}
Method & $\pi_0$ & $\pi_0$ & SpatialVLA & GR00T N1.6 \\
\midrule
Baseline & $88.5 \pm 1.5$ & $36.1 \pm 3.9$ & $43.1 \pm 3.7$ & $59.7 \pm 3.2$ \\
+ Ours & $\mathbf{90.8 \pm 1.2}$ & $\mathbf{43.8 \pm 3.1}$ & $\mathbf{50.3 \pm 2.6}$ & $\mathbf{65.3 \pm 2.2}$ \\
\bottomrule
\end{tabular}
\captionof{table}{Multi-seed evaluation. Success rates (\%) are reported as mean $\pm$ standard deviation across three training seeds.}
\label{tab:multi_seed}
\end{center}

\subsection{DSN Loss-Term Ablation}
\label{app:loss-ablation}

We ablate the two DSN loss terms with SpatialVLA on SimplerEnv, setting the reconstruction or decorrelation loss weight to zero while keeping all other training and evaluation settings identical to the full method. Removing the reconstruction loss reduces the average success rate from 50.3\% to 47.9\%, and removing the decorrelation loss reduces it to 49.0\%, consistent with their roles: reconstruction preserves the backbone feature, while decorrelation keeps the shared and private embeddings apart.

\subsection{Comparison with VICReg}
\label{app:vicreg}

We compare our contrastive loss with VICReg~\citep{bardes2021vicreg} using SpatialVLA on SimplerEnv. We replace only the alignment loss, retaining the frozen EgoHOD encoder, shared/private decomposition, and all other training and evaluation settings.

VICReg improves the average success rate from 43.1\% to 45.2\%, while the contrastive loss achieves 50.3\%. This may be because VICReg regularizes the variance and covariance of paired embeddings, whereas the contrastive loss exploits in-batch positives and negatives for task-discriminative alignment. The observed gap may also reflect limited hyperparameter exploration for VICReg. We leave a more detailed investigation of the mechanisms underlying this difference to future work.

\subsection{Relation to Task Interference}
\label{app:task-interference}

Task interference arises when jointly trained objectives conflict, with one harming the learning of another. Knowledge Insulation studies the adverse effects of action-expert training on pretrained VLMs and combines gradient isolation with discrete-action supervision to improve knowledge transfer~\citep{driess2026knowledge}.

Our diagnosis concerns a single action objective eroding the pretrained semantic structure, which is closer to fine-tuning distorting pretrained features~\citep{kumar2022finetuning,aghajanyan2021r3f}. Within our method, the tension between semantic alignment and action prediction is close to task interference, and the shared/private decomposition is designed to mitigate it structurally.

\subsection{Qualitative Comparisons}
\label{app:qualitative}

Fig.~\ref{fig:qualitative} compares rollouts on the Task generalization axis. Given an unseen source--target combination, the $\pi_0$ baseline selects a suboptimal arm and grasps an irrelevant object. Our method correctly identifies the target and chooses the appropriate arm, demonstrating compositional generalization.

\begin{center}
    \includegraphics[width=\textwidth]{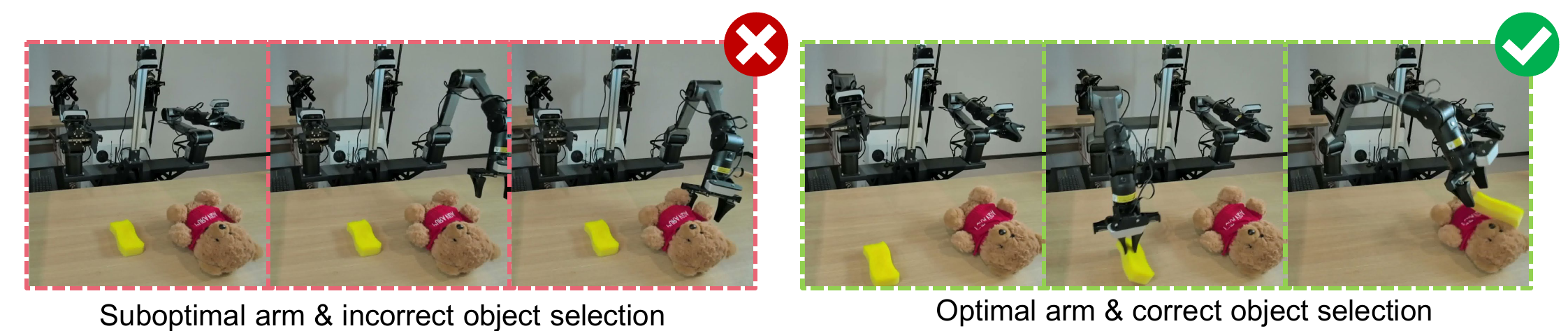}
    \captionof{figure}{Qualitative comparison on the Task axis. \textbf{Left:} $\pi_0$ baseline (failure). \textbf{Right:} Ours (success).}
    \label{fig:qualitative}
\end{center}

\end{document}